\pdfoutput=1
\documentclass[11pt]{article}

\usepackage[final]{acl}

\usepackage{times}
\usepackage{latexsym}

\usepackage[T1]{fontenc}

\usepackage[utf8]{inputenc}

\usepackage{microtype}

\usepackage{inconsolata}

\usepackage{graphicx}

\usepackage{amsfonts}
\usepackage{amsmath}
\usepackage{amssymb}
\usepackage{todonotes}
\usepackage{booktabs}
\usepackage{stfloats}
\usepackage{caption}
\usepackage{subcaption}
\usepackage{hyperref}
\usepackage[table]{xcolor}
\usepackage[most]{tcolorbox}

\title{What's the Catch? Evaluating Temporal Consistency in Vision-Language Models}

\author{Marek Hradil\textsuperscript{1} \quad Danae S\'{a}nchez Villegas\textsuperscript{1,2} \\
        \textsuperscript{1}University of Copenhagen, \textsuperscript{2}The National Center for AI in Society (CAISA) \\
        \texttt{rsc784@alumni.ku.dk}
        \texttt{davi@samf.ku.dk} \\
}

\begin{document}
\maketitle
\begin{abstract}
Vision-language models (VLMs) achieve strong performance on video and image-sequence benchmarks, yet it remains unclear whether they capture temporal structure. To study this question, we formulate temporal grounding as an anomaly detection problem, providing a simple and controlled evaluation that directly tests sensitivity to temporal consistency. We introduce TimeCatch, where temporal anomalies are created by swapping consecutive frames and frame-level anomalies by replacing a frame with Gaussian noise. Models are evaluated on anomaly detection and localization tasks across four synthetic and real-world datasets, alongside a human study.
Our evaluation reveals a substantial gap between frame-level and temporal anomaly detection. While VLMs consistently detect frame-level anomalies and often localize them accurately, under our main evaluation setting they generally perform near chance on temporal anomaly detection and show limited localization performance. Humans, in contrast, achieve near-ceiling performance on both tasks. Additional analyses across model scales, prompting strategies, sequence lengths, and visual similarity show that performance can improve under some conditions, while substantial gaps in temporal anomaly detection and localization remain.
Together, these findings reveal a gap between frame-level and temporal anomaly detection. TimeCatch provides a controlled benchmark for evaluating temporal consistency in vision-language models.\footnote{Code and the results are hosted at: \url{https://github.com/marek-hradil/timecatch}}
\end{abstract}

\section{Introduction}


\begin{figure*}[t]
    \centering
    \includegraphics[width=1\linewidth]{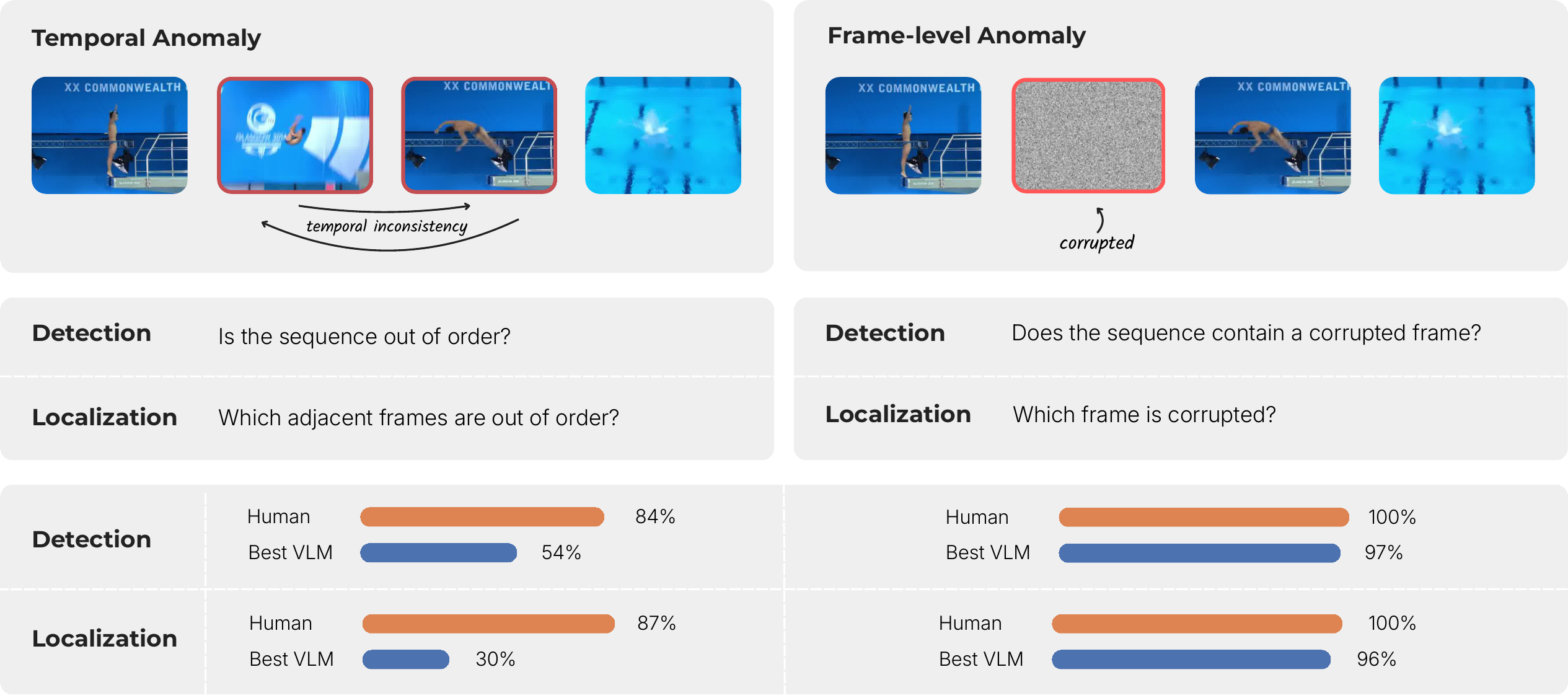}
    \caption{Overview of the proposed anomaly detection tasks. Given an image sequence, models detect or localize either temporal anomalies (swapped adjacent frames) or frame-level anomalies (corrupted frames). Under our main evaluation setting, VLMs perform well on frame-level tasks but generally remain close to chance on temporal anomaly tasks, whereas humans achieve high performance on both.}
    \label{fig:visual-abstract}
\end{figure*}

Vision-language models (VLMs) are increasingly capable of processing videos and image sequences, achieving strong performance across a wide range of multimodal benchmarks \cite{zhang2023video,zhu2024minigpt,zhang2024mm,li2025videochat, bai2025qwen3vltechnicalreport}. As a result, they are being deployed in settings where decisions depend not only on what is visible in individual frames, but also on how visual information evolves over time. Examples include autonomous driving \cite{zhou2024vision}, robotics \cite{kawaharazuka2025vision}, medical imaging \cite{van2024large}, and video understanding, where recognizing temporal consistency is often as important as recognizing objects or actions themselves. Despite this progress, it remains unclear to what extent current VLMs capture temporal structure \cite{11491967}. Existing benchmarks frequently rely on question answering or captioning tasks, making it difficult to determine whether successful performance reflects temporal reasoning or the exploitation of shortcuts \cite{cores2025losttimenewtemporal,xue2026seeing}. Prior work has shown that many video understanding benchmarks can be solved using only a small subset of frames \cite{buch2022revisiting,lei2023revealing,krojer2025shortcut}, textual biases \cite{goyal2017making}, or are largely invariant to changes in event ordering \cite{cores2025losttimenewtemporal,xue2026seeing}. Consequently, strong benchmark performance does not necessarily indicate that a model captures temporal structure or can detect violations of temporal consistency in visual sequences.

To study this question, we focus on a fundamental capability: recognizing when a visual sequence is temporally inconsistent. Detecting such inconsistencies provides a controlled probe of temporal reasoning, as the task requires judging only whether the temporal evolution of a sequence is plausible. We distinguish between two types of anomalies: (i) \emph{temporal anomalies}, which arise only through inconsistencies across multiple frames and require reasoning about how events unfold over time; and (ii) \emph{frame-level anomalies}, which occur within individual frames and can be identified without temporal context, serving as a control condition that disentangles temporal reasoning from frame-level reasoning. To this end, we introduce \textbf{TimeCatch}, a benchmark for evaluating temporal grounding through anomaly detection in image sequences. Temporal anomalies are generated by swapping consecutive frames within a sequence, while frame-level anomalies are generated by replacing a frame with Gaussian noise. Figure~\ref{fig:visual-abstract} provides an overview of the proposed benchmark. Models are evaluated on both anomaly detection and anomaly localization tasks across four datasets spanning synthetic and real-world domains.
Our experiments reveal a substantial performance gap between frame-level and temporal anomaly detection. While VLMs reliably detect frame-level anomalies and often localize them accurately, in the main evaluation setting they generally perform near chance on temporal anomaly detection and achieve limited performance on temporal anomaly localization. Humans, by contrast, achieve near-ceiling performance on both tasks. Further analyses across model scales, prompts, sequence lengths, and visual similarity suggest that these failures cannot be explained solely by an inability to perceive the visual sequence.
Overall, our findings reveal a persistent gap between frame-level anomaly performance and sensitivity to temporal consistency. 
TimeCatch provides a simple and controlled way to evaluate this capability. Our contributions are:

\begin{itemize}
    \item A systematic evaluation of state-of-the-art VLMs and human performance on temporal anomaly detection and localization.

    \item A controlled comparison of VLM and human performance on temporal and frame-level anomaly tasks, revealing a substantial gap in sensitivity to temporal consistency. 

    \item \textbf{TimeCatch}, a controlled benchmark for evaluating temporal grounding through temporal and frame-level anomaly detection.
\end{itemize}

\section{Related Work}
\subsection{Temporal Reasoning Benchmarks}

Recent work has proposed a variety of benchmarks for evaluating temporal reasoning in vision-language models. These can broadly be grouped into three categories. Question answering benchmarks, such as TGIF-QA~\cite{jang2017tgif}, TempCompass~\cite{liu2024tempcompass}, VidHalluc~\cite{li2025vidhalluc}, TVBench~\cite{cores2025losttimenewtemporal}, and Mementos~\cite{wang2024mementos}, assess temporal understanding through multiple-choice or free-form responses. Discriminative benchmarks, including Vinoground~\cite{zhang2024vinoground} and MVP~\cite{krojer2025shortcut}, require models to distinguish between candidate videos or descriptions.  
Finally, ordering benchmarks, such as TimeBlind~\cite{li2026timeblind}, TOMATO~\cite{shangguan2025tomato}, and AoTBench~\cite{xue2026seeing}, evaluate whether models can reason about event order.

While question answering and discriminative benchmarks provide indirect measures of temporal reasoning, successful performance does not necessarily require identifying violations of temporal consistency. Ordering-based and anomaly-detection approaches are more closely related to our setting. In particular, TempVS~\cite{song2025burn} formulates temporal reasoning as an image-ordering task over a single composite image containing multiple frames. 
Prior benchmarks, including TempVS, MVP, and TimeBlind, have reported substantial human–model performance gaps on tasks requiring temporal reasoning. VANE-Bench~\cite{gani-etal-2025-vane} evaluates Video-LMMs on the detection and localization of anomalies and inconsistencies in videos. In contrast, TimeCatch specifically isolates temporal consistency by introducing anomalies through frame reordering while leaving the individual frames unchanged.

\begin{figure*}[t!]
    \centering
    \includegraphics[width=1\linewidth, trim={1.2cm 0 1.2cm 0},clip]{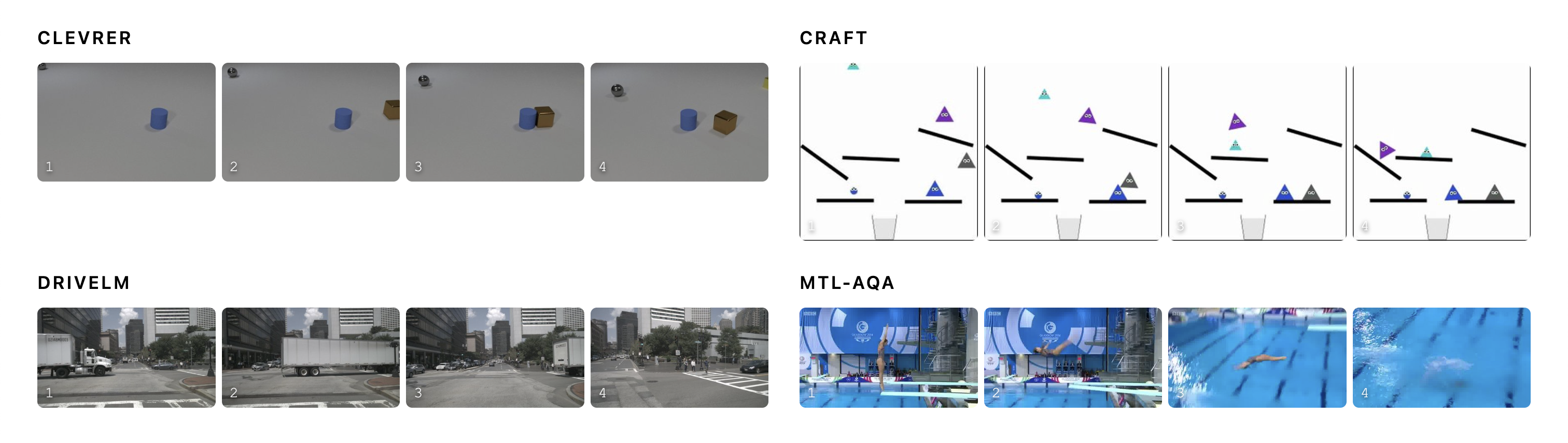}
    \caption{Example image sequences from the four benchmark datasets. The benchmark spans both synthetic (CLEVRER, CRAFT) and real-world (DriveLM, MTL-AQA) domains.
    }
    \label{fig:dataset-examples}
\end{figure*}

\subsection{Challenges in Evaluating Temporal Reasoning}

Strong performance on video understanding benchmarks does not necessarily imply robust temporal reasoning. A growing body of work has shown that models can exploit shortcuts embedded in benchmark design to achieve high accuracy without processing the full temporal content of a sequence. Several studies demonstrate that sequence-level questions can often be answered using only a small subset of frames~\cite{buch2022revisiting,lei2023revealing}, while \citet{krojer2025shortcut} show that VLMs frequently skip large portions of image sequences while maintaining performance. In addition to visual shortcuts, models may exploit biases in task formulations. Prior work has shown that textual cues can strongly influence model predictions~\cite{goyal2017making,villegas2026reasoningdynamicslimitsmonitoring}, and similar effects have been observed in modern video benchmarks~\cite{cores2025losttimenewtemporal}. Recently, \citet{xue2026seeing} and \citet{cores2025losttimenewtemporal} demonstrate that performance on several temporal benchmarks remains largely unchanged when event order is shuffled, suggesting that temporal ordering is often not required to solve the task. 

These findings highlight two challenges for evaluating temporal reasoning: isolating temporal consistency from other sources of information and minimizing opportunities for shortcut exploitation. TimeCatch addresses both by introducing controlled temporal perturbations while keeping the underlying visual content unchanged, allowing temporal consistency to be evaluated in a controlled setting.

\section{Evaluation Framework}

We introduce \textbf{TimeCatch}, a benchmark for evaluating temporal grounding through anomaly detection in image sequences. Our goal is to evaluate an aspect of temporal grounding, specifically, sensitivity to temporal consistency, in a controlled setting that minimizes reliance on language biases and does not require domain expertise. TimeCatch consists of four tasks:
(i) temporal anomaly detection,
(ii) temporal anomaly localization,
(iii) frame-level anomaly detection,
(iv) frame-level anomaly localization.
These tasks do not require forecasting future states or domain-specific knowledge; they require only recognizing that the observed sequence is inconsistent with a plausible temporal progression.

\subsection{Temporal Anomaly Tasks}

\paragraph{Temporal Anomaly Detection (Temporal Detect)}
Given an image sequence $S=(I_1,\ldots,I_n)$, a temporal anomaly is introduced by swapping a randomly selected consecutive frame pair $(I_i, I_{i+1})$, where $i \sim \mathcal{U}\{1,\ldots,n-1\}$. The model is tasked with predicting whether the resulting sequence contains a temporal anomaly.

\paragraph{Temporal Anomaly Localization (Temporal Localize)}
Using the same anomaly generation procedure, the model is given a sequence containing a guaranteed frame swap and must predict the swap location $i$.

\subsection{Frame-Level Anomaly Tasks}

To disentangle temporal reasoning from anomaly perception, we introduce a control condition based on frame-level anomalies by replacing a frame with Gaussian noise. Gaussian noise provides an unambiguous frame-level anomaly without altering the sequence structure or introducing semantic content. 

Strong performance on these tasks demonstrates that a model can attend to the sequence and identify salient anomalous frames, suggesting that failures on temporal anomalies more specifically reflect limitations in temporal reasoning.

\paragraph{Frame-Level Anomaly Detection (Frame Detect)}
Given an image sequence $S=(I_1,\ldots,I_n)$, a frame-level anomaly is introduced by replacing a uniformly sampled frame $I_i$, where $i \sim \mathcal{U}{1,\ldots,n}$, with Gaussian noise. The model is tasked with predicting whether the resulting sequence contains a frame-level anomaly.

\paragraph{Frame-Level Anomaly Localization (Frame Localize)}
Using the same anomaly generation procedure, the model is given a sequence containing a guaranteed corrupted frame and must predict the anomaly location $i$.

\subsection{Dataset Curation}

\begin{figure*}[t!]
    \centering
    \includegraphics[width=1\linewidth]{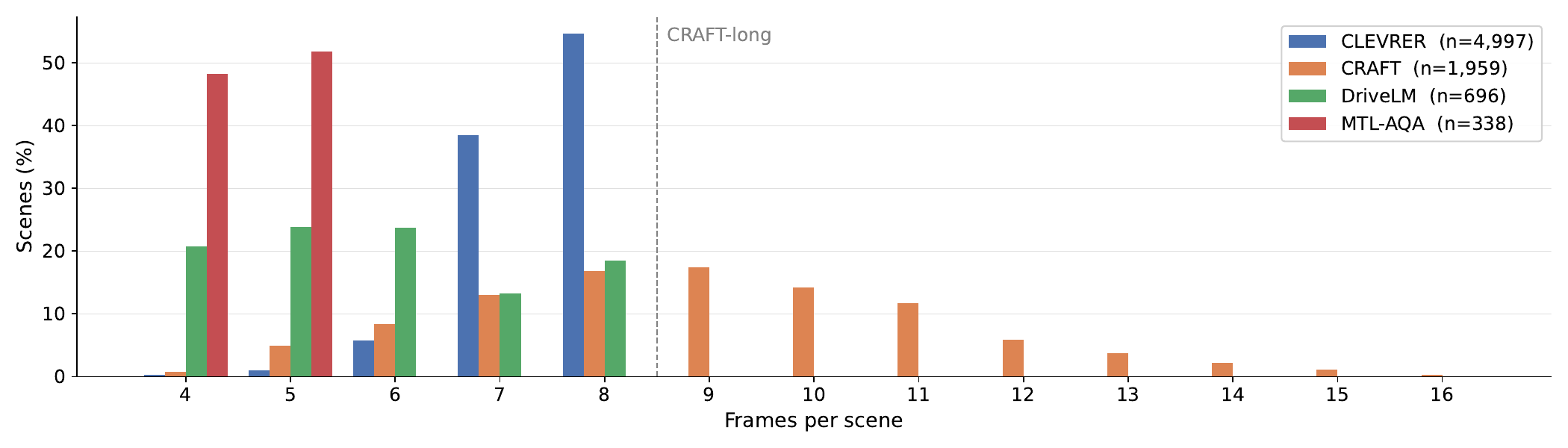}
    \caption{Sequence length distribution across the benchmark datasets. Sequences of 4 to 8 frames are used in the main experiments. Longer CRAFT sequences (9--16 frames) are reserved for evaluating the effect of sequence length on temporal anomaly detection.}
    \label{fig:dataset-histograms}
\end{figure*}

\begin{table}[t!]
	\centering
	\small
	\begin{tabular}{lcccc}
		\toprule
		Dataset    & Domain     & Sequences & Avg.\ Length \\
		\midrule
		CLEVRER    & Synthetic  & 4,997     & 7.5          \\
		CRAFT      & Synthetic  & 858       & 6.9          \\
		DriveLM    & Real-world & 696       & 5.9          \\
		MTL-AQA    & Real-world & 338       & 4.5          \\
		CRAFT-Long & Synthetic  & 1,101     & 10.6         \\
		\bottomrule
	\end{tabular}
	\caption{Statistics of datasets included in TimeCatch.}
	\label{tab:dataset_stats_1}
\end{table}

To construct a controlled benchmark, we select datasets that satisfy four criteria: (i) events should be interpretable without specialized domain knowledge, (ii) the sequence should follow a single temporal trajectory, (iii) consecutive frames should exhibit distinguishable changes, and (iv) the viewpoint should remain consistent throughout the sequence. These requirements help isolate temporal consistency from confounding factors such as scene cuts, domain expertise, or ambiguous event structure. The benchmark draws on four dataset test splits, spanning both synthetic and real-world domains. 
\paragraph{Synthetic Datasets} CLEVRER~\cite{yi2020clevrer} and CRAFT~\cite{ates2022craft} contain object interactions in 3D and 2D environments, respectively. 

\paragraph{Real-world Datasets} DriveLM~\cite{sima2024drivelm} provides image sequences of real driving scenarios, while MTL-AQA~\cite{parmar2019and} contains competitive diving videos. Figure~\ref{fig:dataset-examples} provides an example for each dataset.

\paragraph{Sampling and Filtering}
For the video datasets (CRAFT, CLEVRER, and MTL-AQA), we convert videos into image sequences through temporal subsampling. Frames are sampled every 1.5 seconds for CRAFT and CLEVRER, and every 1 second for MTL-AQA, reflecting the different rates at which visually distinguishable changes occur in the underlying videos.
To ensure that temporal anomalies remain perceptually meaningful, we further filter sequences using LPIPS~\cite{zhang2018unreasonable}, a metric that correlates with human judgments of visual similarity. Consecutive frames whose LPIPS distance falls below 0.05 are removed, as such pairs often exhibit little observable change and make temporal anomalies difficult to perceive. Sequences containing fewer than four frames after filtering are discarded, as very short sequences provide limited temporal context and few possible anomaly locations. Figure~\ref{fig:dataset-histograms} shows the sequence length distribution across the benchmark datasets. The majority of sequences contain 4--8 frames and are used in the main evaluation. Longer sequences from CRAFT (9--16 frames) are treated as a separate subset, \textit{CRAFT-long}, and reserved for the sequence length analysis in Section~\ref{sec:sec_length}. Table~\ref{tab:dataset_stats_1} summarizes the datasets included in TimeCatch.

\paragraph{Scene Descriptions} Finally, we construct scene descriptions from the available dataset annotations (see Appendix~\ref{appendix:scene_description} for details), providing models with high-level semantic context while preserving the temporal nature of the task.

\section{Experimental Setup}

\subsection{Models}

We evaluate five open-weight vision-language models: Qwen2.5-VL-7B~\cite{bai2025qwen25vltechnicalreport}, Qwen3-VL-8B~\cite{bai2025qwen3vltechnicalreport}, Gemma-4-E4B~\cite{gemma4modelcard},  InternVL3-8B~\cite{zhu2025internvl3}, InternVL3.5-8B~\cite{wang2025internvl3}. These models represent recent state-of-the-art VLMs with native support for multi-image inputs. We focus on open-weight models to ensure reproducibility and enable controlled inference with consistent model configurations and decoding procedures across experiments.

\begin{table*}[t!]
    \centering
    \setlength{\tabcolsep}{4pt}
    \small
    \begin{tabular}{l cccc cccc cccc cccc}
    \toprule
     & \multicolumn{4}{c}{Temporal Detect} & \multicolumn{4}{c}{Temporal Localize} & \multicolumn{4}{c}{Frame Detect} & \multicolumn{4}{c}{Frame Localize} \\
    \cmidrule(lr){2-5} \cmidrule(lr){6-9} \cmidrule(lr){10-13} \cmidrule(lr){14-17}
     & CL & CR & DR & MT & CL & CR & DR & MT & CL & CR & DR & MT & CL & CR & DR & MT \\
    \midrule
     Random & 50.0 & 50.0 & 50.0 & 50.0 & 4.3 & 5.5 & 8.7 & 13.2 & 50.0 & 50.0 & 50.0 & 50.0 & 13.5 & 14.9 & 18.1 & 22.4 \\
     \midrule
    Qwen2.5-VL-7B & 50.5 & 47.6 & 50.7 & 45.3 & 13.1 & 14.3 & 14.8 & 29.9 & 66.9 & 67.9 & 60.3 & 71.6 & 42.0 & 60.5 & 45.8 & 73.1 \\
    Qwen3-VL-8B & \textbf{51.9} & \textbf{53.3} & 51.9 & \textbf{57.4} & \textbf{25.6} & \textbf{24.5} & \textbf{24.7} & \textbf{43.2} & \textbf{93.3} & \textbf{98.1} & \textbf{99.6} & \textbf{94.1} & \textbf{98.9} & \textbf{96.3} & \textbf{92.7} & \textbf{99.4} \\
    Gemma-4-E4B & 49.2 & 48.0 & \textbf{53.9} & 56.2 & 15.0 & 14.7 & 22.3 & 29.9 & 77.2 & 80.5 & 82.0 & 76.6 & 33.6 & 46.5 & 61.9 & 88.2 \\
    InternVL3-8B & 50.1 & 49.1 & 53.2 & 47.9 & 14.3 & 17.7 & 17.8 & 21.6 & 58.1 & 77.7 & 68.4 & 76.6 & 30.9 & 52.9 & 33.9 & 65.7 \\
    InternVL3.5-8B & 49.6 & 49.9 & 49.9 & 50.6 & 15.4 & 18.2 & 9.8 & 33.1 & 66.9 & 66.7 & 69.7 & 77.2 & 18.6 & 22.4 & 16.4 & 29.6 \\
    \bottomrule
    \end{tabular}
    \caption{Detection and localization accuracy (\%) across datasets. Models reliably detect and localize frame-level anomalies but struggle on temporal anomaly tasks. Results with 95\% Wilson score confidence intervals are reported in Table~\ref{tab:detailed_results_ci} Appendix~\ref{appendix:additional-results}. CL: CLEVRER, CR: CRAFT, DR: DriveLM, MT: MTL-AQA.}
    \label{tab:detailed_results}
\end{table*}

\subsection{Evaluation Protocol}

\paragraph{Metric}
We evaluate both anomaly detection and anomaly localization using accuracy. For detection, accuracy is appropriate because the classes are balanced by construction. For localization, accuracy measures the fraction of sequences for which the anomaly position is correctly identified.

\paragraph{Implementation Details}
All models are evaluated using a unified prompting protocol described in Appendix~\ref{appendix:prompting}. Models are served using vLLM~\cite{kwon2023efficient}. To ensure consistent evaluation, constrained decoding is applied throughout: detection outputs are restricted to \texttt{\{yes,no\}}, while localization outputs are restricted to valid frame indices. All experiments are conducted on NVIDIA A100 GPUs. All models are evaluated in a zero-shot setting.

\section{Results}

\subsection{Main Results}

\paragraph{Temporal Anomaly Detection} Table~\ref{tab:detailed_results} reports temporal anomaly detection and localization performance across all evaluated models. In the main evaluation setting reported in Table 2, the highest temporal detection accuracy is 57.4\%, and the highest temporal localization accuracy is 43.2\%. Overall, detection performance is generally close to chance, while localization remains limited across models and datasets. 

\paragraph{Frame-Level Anomaly Detection}

\begin{figure*}[t!]
    \centering
    \includegraphics[width=1\linewidth]{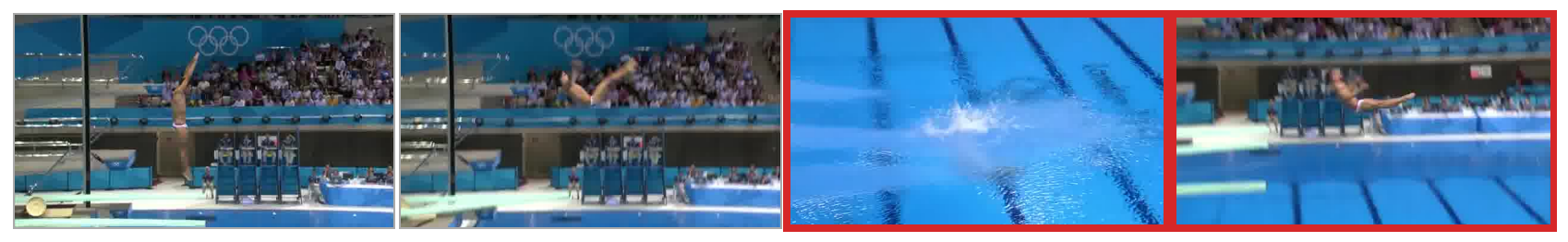}
    \caption{Example from the temporal anomaly localization task on MTL-AQA. The highlighted frames indicate the swapped pair. Although the temporal inconsistency is readily identified by the human participant, all evaluated VLMs fail to localize the anomaly correctly.}
    \label{fig:model-failure}
\end{figure*}

\begin{figure}[t!]
    \centering
    \includegraphics[width=1\linewidth]{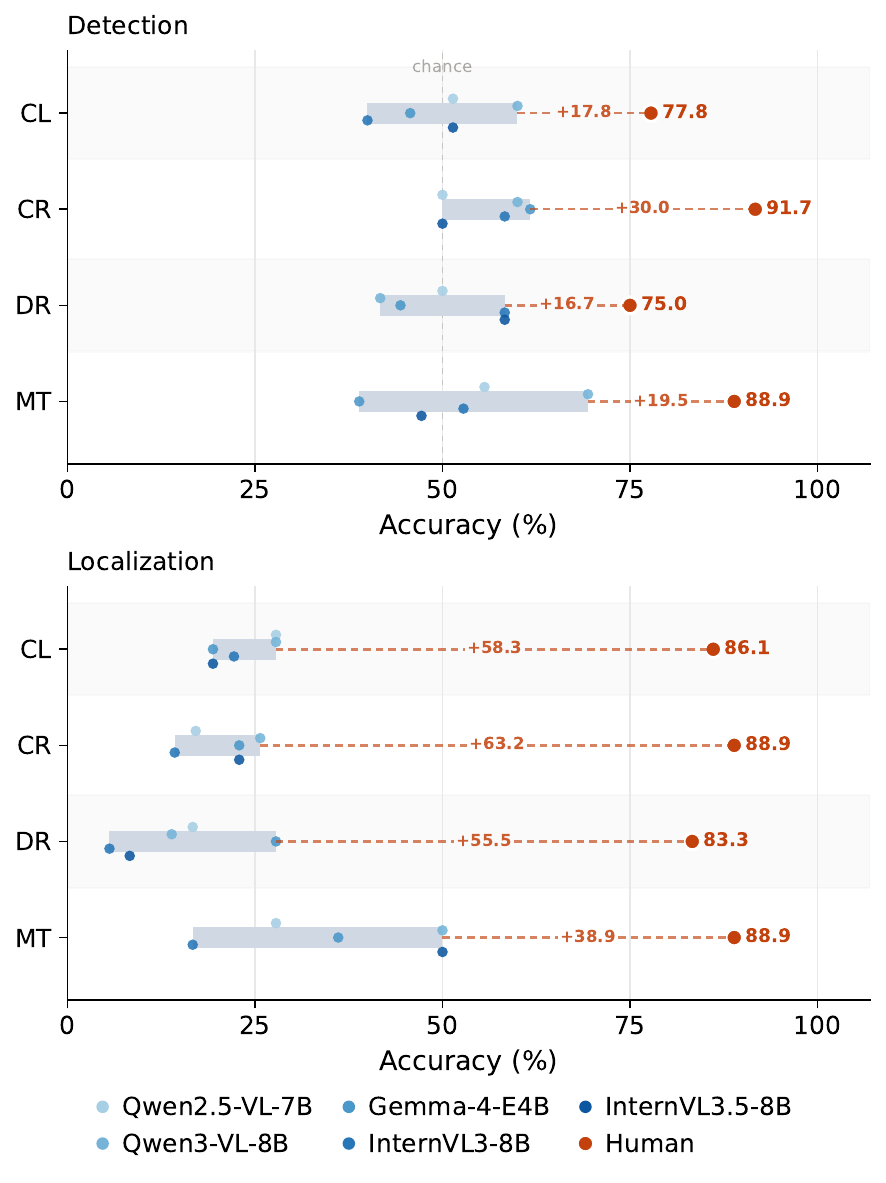}
    \caption{Temporal anomaly detection and localization accuracy (\%) on the human study subset. Humans consistently outperform all evaluated VLMs across datasets, highlighting a substantial gap between human and VLM performance on temporal reasoning. We also perform statistical significance tests for this experiment, reported in Table~\ref{tab:mcnemar_pooled} Appendix~\ref{appendix:additional-results}.}
    \label{fig:human_results}
\end{figure}

In contrast, models achieve substantially higher performance on frame-level anomaly tasks. Detection and localization accuracy are consistently high across datasets, reaching up to 99.6\% and 99.4\%, respectively. Success on these control tasks demonstrates that VLMs can attend to image sequences and identify anomalous frames. The large gap between frame-level and temporal anomalies therefore suggests that temporal consistency poses an additional challenge beyond detecting salient anomalies within individual frames. 

\subsection{Human Study}

To establish a human reference point, we conduct a human study on both \emph{temporal} anomaly detection and localization. Participants complete the same tasks as the evaluated models using a custom annotation platform. Annotation instructions, interface screenshots, and additional details of the study protocol are provided in Appendix~\ref{appendix:human_study}.

Figure~\ref{fig:human_results} compares human and model performance on a subset of 72 samples from each dataset. Across datasets, humans achieve 75.0--91.7\% accuracy on temporal anomaly detection and 83.3--88.9\% on temporal anomaly localization, substantially outperforming all evaluated VLMs. 
On this shared human-study subset, no evaluated model exceeds 70\% detection accuracy or 50\% localization accuracy on any dataset. This indicates that the temporal anomalies used in our evaluation are generally detectable by humans and suggests that the lower VLM performance cannot be attributed solely to inherent task difficulty.
Figure~\ref{fig:model-failure} illustrates a representative failure case.\footnote{More qualitative examples in Figure \ref{fig:quadrant-examples} Appendix \ref{appendix:additional-results}.} Although each frame appears plausible in isolation, identifying the anomaly requires reasoning about the temporal progression of the event rather than detecting abnormalities within individual frames. While the human participant correctly localizes the swapped frames, all evaluated VLMs fail on this example.

\begin{figure}[t!]
    \centering
    \includegraphics[width=1\linewidth]{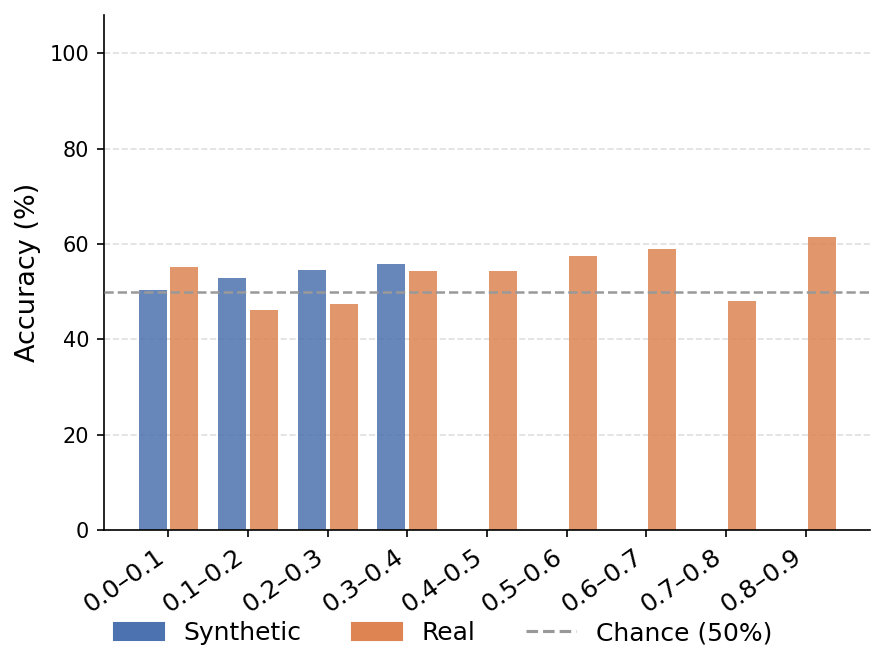}
    \caption{
    Temporal anomaly detection accuracy by LPIPS distance for Qwen3-VL-8B. Although larger perceptual differences between swapped frames slightly improve performance, accuracy remains close to chance across both synthetic and real-world datasets, indicating that visual similarity is not the primary limitation.}
    \label{fig:visual-similarity}
\end{figure}

\section{Analysis}
\label{sec:analysis}

Since Qwen3-VL-8B achieves the best overall performance on TimeCatch, we use the Qwen3 family for the following analyses.

\begin{figure}
    \centering
    \includegraphics[width=0.75\linewidth]{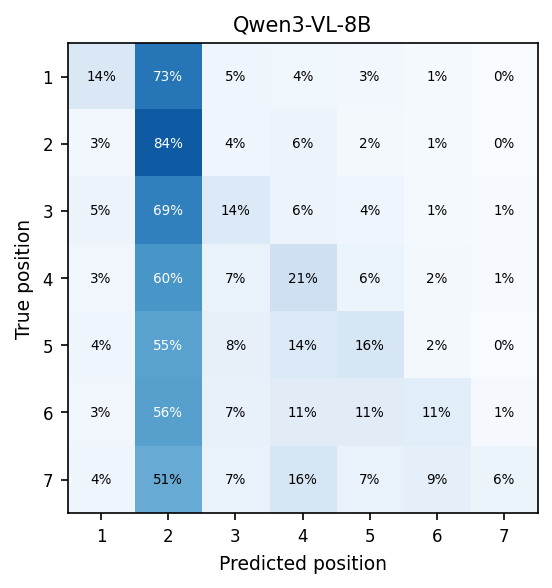}
    \caption{Temporal anomaly localization confusion matrix for Qwen3-VL-8B, averaged across all datasets. Rather than correctly localizing the anomaly, the model exhibits a strong bias toward predicting position 2, independent of the true anomaly location.}
    \label{fig:localize-heatmap-qwen}
\end{figure}

\begin{figure}[t!]
    \centering
    \includegraphics[width=1\linewidth]{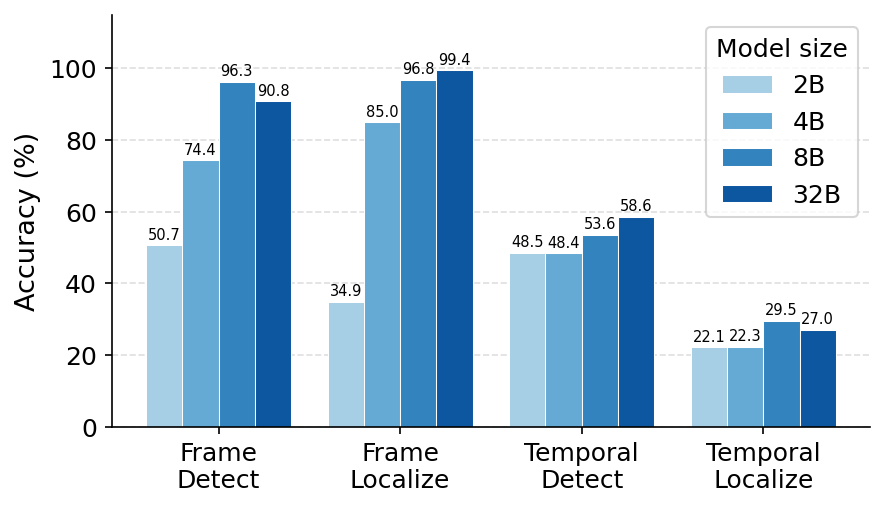}
    \caption{Effect of model scale on anomaly detection and localization accuracy for the Qwen3-VL family. Scaling improves frame-level anomaly performance and also yields gains in temporal anomaly detection, particularly from 4B to 32B, while temporal localization remains comparatively low. This suggests that model capacity contributes to temporal anomaly detection, but does not close the performance gap observed on temporal tasks.}
    \label{fig:scaling}
\end{figure}

\begin{figure}[t!]
    \centering
    \includegraphics[width=1\linewidth]{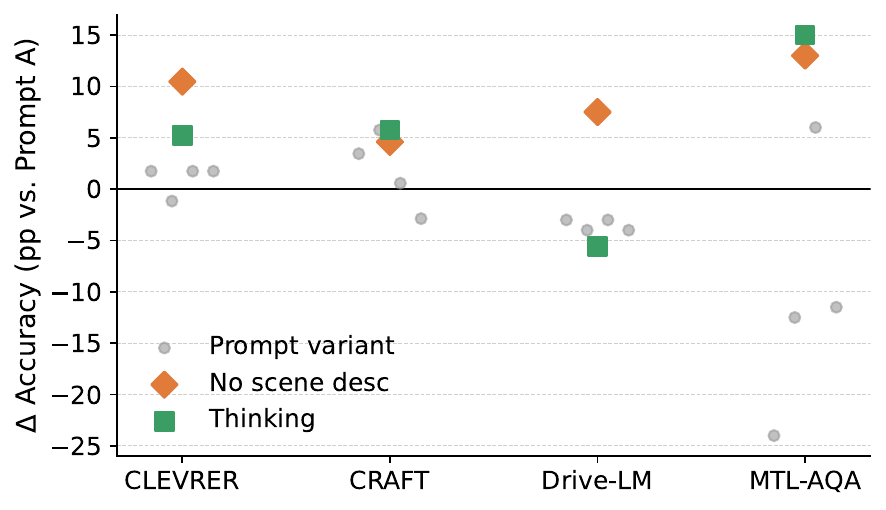}
    \caption{Effect of prompting on temporal anomaly detection accuracy. Values show the change in accuracy (percentage points) relative to the base prompt (Prompt A) for Qwen3-VL-8B. Grey circles correspond to alternative prompt formulations, while orange diamonds and green squares denote removing scene descriptions and enabling reasoning, respectively. 
    }
    \label{fig:textual-grounding-deltas}
\end{figure}

\begin{figure*}[t!]
    \centering
    \includegraphics[width=1\linewidth]{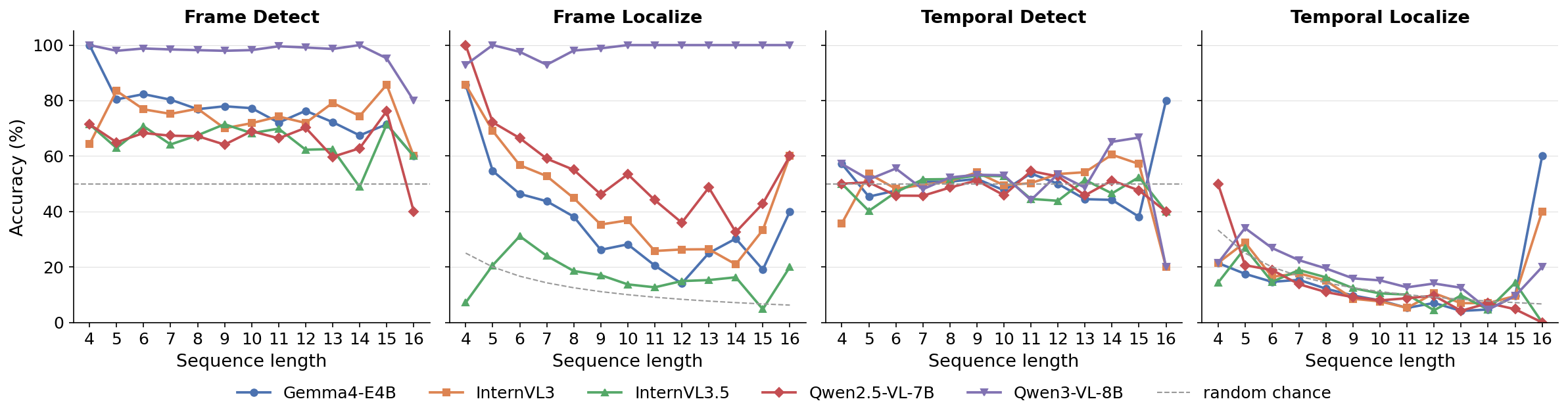}
    \caption{Effect of sequence length on anomaly detection and localization accuracy across sequence lengths (4--16 frames). Frame-level detection remains high, whereas frame-level localization degrades as the number of frames increases. Temporal anomaly detection and localization show little dependence on sequence length. The final bucket ($n=16$) contains only five samples and should be interpreted with caution (see Figure \ref{fig:dataset-histograms}).}
    \label{fig:by-length}
\end{figure*}

\subsection{Does Visual Similarity Explain the Failures?} 
To investigate whether temporal anomaly detection depends on low-level perceptual cues, we group swapped frame pairs according to their LPIPS distance and measure detection accuracy within each bin. Each bin covers a fixed interval of 0.1 LPIPS, such that pairs in the first bin ($0.0$--$0.1$) are highly similar, while pairs in the last bin ($0.8$--$0.9$) are perceptually distinct. Figure~\ref{fig:visual-similarity} shows results for Qwen3-VL-8B. While performance increases slightly for more perceptually distinct frame pairs, the gains are modest and accuracy remains far below human levels across all bins. 
Overall, visual similarity has only a limited effect on temporal anomaly detection. Full per-model results are provided in Appendix~\ref{appendix:additional-results}.

\subsection{Do Incorrect Predictions Occur Near the True Location?}

While localization accuracy measures exact correctness, it does not capture whether incorrect predictions occur near the true anomaly location. We therefore analyze the distribution of predicted positions relative to ground truth using confusion matrices. Figure~\ref{fig:localize-heatmap-qwen} shows the results for Qwen3-VL. If the model were identifying the relevant region of the sequence but failing to localize the anomaly precisely, predictions would cluster around the diagonal. Instead, the model exhibits a strong bias toward predicting position 2 regardless of the true anomaly location, suggesting that localization errors arise from a systematic prediction bias rather than near misses. Confusion matrices for all models are provided in Appendix~\ref{appendix:additional-results}.

\subsection{Does Increasing Model Size Improve Performance?}

We investigate whether temporal anomaly detection improves with model capacity by evaluating Qwen3-VL across four model sizes (2B, 4B, 8B, and 32B). Figure~\ref{fig:scaling} shows that performance on frame-level anomaly tasks improves consistently with model size, reaching near-ceiling accuracy for the largest models. Temporal anomaly detection also improves with scale, particularly from 4B to 32B, although the gains are smaller than for frame-level detection. Temporal localization shows less consistent improvement and remains comparatively low. These results suggest that increased model capacity benefits temporal anomaly detection, but does not fully address the difficulty of reasoning about temporal consistency.

\subsection{Can Prompting Improve Temporal Reasoning?}

To investigate whether the observed failures depend on textual context or prompting, we evaluate multiple prompt formulations, remove scene descriptions, and compare standard and reasoning-enabled (thinking) models. As shown in Figure~\ref{fig:textual-grounding-deltas}, alternative prompt formulations have little effect on temporal anomaly detection. Removing scene descriptions often yields modest improvements, suggesting that the accompanying text may distract models from reasoning about temporal consistency. By contrast, enabling reasoning produces mixed results, improving performance on some datasets while degrading it on others, with no consistent overall benefit. Table~\ref{tab:no_scene_desc_delta} in Appendix~\ref{appendix:additional-results} reports the per-model change in accuracy after removing scene descriptions and shows that the largest improvements occur on the frame-level tasks rather than the temporal ones. Overall, prompt variations yield no consistent improvement on temporal tasks. 

\subsection{Does Increasing Sequence Length Improve Temporal Reasoning?}
\label{sec:sec_length}
To investigate whether longer image sequences improve performance, we evaluate models on longer CRAFT sequences containing up to 16 frames. Figure~\ref{fig:by-length} shows that temporal anomaly performance remains near chance across sequence lengths. Increasing the number of frames does not improve accuracy, indicating that temporal anomaly performance remains largely unchanged despite additional context. In contrast, frame-level detection remains consistently high across sequence lengths, while frame-level localization degrades as the number of candidate frames increases, suggesting that precisely identifying the corrupted frame becomes more challenging in longer sequences.

\section{Implications for Temporal Grounding}

Our findings support concerns that existing video benchmarks may overestimate temporal reasoning capabilities. A model may successfully identify objects, actions, or anomalous single frames while remaining insensitive to the temporal relationships between them. By modifying only the order of frames and leaving the visual content unchanged, TimeCatch isolates this aspect of temporal grounding and provides a direct test of sensitivity to temporal consistency. 
Overall, temporal anomaly performance remains substantially weaker than frame-level anomaly performance across the visual-similarity, prompting, model-scale, and sequence-length conditions we examine, although some conditions, specifically, increased model scale and removing scene descriptions on particular datasets, yield improvements. This suggests that the observed difficulty is not specific to a single evaluation configuration, while also indicating that model capacity and prompting context can influence performance.
These findings may also have implications for applications where decisions depend on how events unfold over time. In domains such as autonomous driving, robotics, and video understanding, recognizing whether image sequences are temporally consistent can be as important as recognizing the observations themselves. We therefore view temporal anomaly detection as a complementary evaluation setting for measuring progress toward temporally grounded VLMs.

\section{Conclusion}

In this work, we conducted a systematic evaluation of state-of-the-art VLMs and human performance on temporal anomaly detection and localization. Our findings reveal a substantial gap between human and VLM performance: while VLMs reliably detect and localize frame-level anomalies, they struggle to recognize temporal inconsistencies, whereas humans achieve near-ceiling performance on both temporal tasks. To enable this evaluation, we introduced TimeCatch, a controlled benchmark for evaluating temporal grounding through temporal and frame-level anomaly detection. We hope TimeCatch will serve as a complementary benchmark for tracking progress in temporal reasoning. Future work could extend the benchmark with other types of frame-level controls, such as plausible distractor frames, as well as motion continuity and causal event structure evaluation.

\section*{Limitations}

TimeCatch evaluates a specific aspect of temporal grounding: sensitivity to violations of temporal consistency introduced through adjacent frame swaps. While this provides a controlled setting for evaluating sensitivity to temporal consistency, it does not capture all forms of temporal reasoning required in real-world video understanding, such as long-range dependencies, causal event reasoning, or continuity of motion. We therefore view TimeCatch as a complementary diagnostic benchmark that isolates a fundamental capability, rather than a comprehensive evaluation of temporal grounding.

Our frame-level control uses Gaussian noise as a test of whether models can identify and localize anomalous information within an image sequence, rather than as a difficulty-matched comparison with the temporal task. The magnitude of the frame-level–temporal performance gap may therefore partly reflect differences in anomaly salience.

\section*{Ethical Considerations}

TimeCatch is intended as a diagnostic benchmark rather than a capability-enhancing method: it evaluates a specific limitation in VLMs' temporal reasoning rather than enabling new functionality, and we see no direct misuse risk. We use only publicly available datasets released for research use and adhere to their respective licenses and terms of use. Further details on data licensing and human study procedures are provided in Appendix~\ref{appendix_dataset_details} and Appendix~\ref{appendix:human_study} respectively.

\section*{Acknowledgments}
This work was supported by a research grant (VIL53122) from VILLUM FONDEN. We acknowledge the EuroHPC Joint Undertaking for awarding this project access to the EuroHPC supercomputer MareNostrum 5, hosted by the Barcelona Supercomputing Center (BSC), Spain, under project ID EHPC-DEV-2025D11-030.

\bibliography{custom}

\newpage
\appendix

\section{Prompting}
\label{appendix:prompting}

\subsection{Main Task Prompts}
The following prompts are used for the main experiments across all datasets.

\begin{tcolorbox}[colback=gray!5,colframe=gray!40,title=Temporal Detect]
You are given a sequence of images showing a scene unfolding over
time. The frames should appear in a natural temporal order, but two
consecutive frames may have been swapped. Reply with only `yes' if you
detect a swap, or `no' if the order looks correct.
\end{tcolorbox}

\begin{tcolorbox}[colback=gray!5,colframe=gray!40,title=Temporal Localize]
You are given a sequence of images showing a scene unfolding over time. Exactly two consecutive frames have been swapped. Reply with only the two frame numbers that are out of order, separated by a comma. For example: \texttt{3,4} means frames 3 and 4 were swapped. Use 1-based indexing (1 for the first frame).
\end{tcolorbox}

\begin{tcolorbox}[colback=gray!5,colframe=gray!40,title=Frame Detect]
You are given a sequence of images showing a scene unfolding over time. One frame may have been replaced with random noise. Reply with only `yes' if you see a corrupted frame, or `no' if all frames look normal.
\end{tcolorbox}

\begin{tcolorbox}[colback=gray!5,colframe=gray!40,title=Frame Localize]
You are given a sequence of images showing a scene unfolding over time. Exactly one frame has been replaced with random noise. Reply with only the number of the corrupted frame. Use 1-based indexing (1 for the first frame).
\end{tcolorbox}

\subsection{Prompt Variations}
We evaluate four different prompt phrasings for the temporal detection task to assess sensitivity to instruction prompt. All variants instruct the model to reply with \texttt{yes} or \texttt{no}.

\paragraph{Prompt B}
\begin{quote}
\ttfamily
You are given a sequence of images showing a scene unfolding over
time. Does the sequence appear to be in the correct temporal order, with
no frames swapped? Reply with only `yes' if the order is correct, or
`no' if something looks wrong.
\end{quote}

\paragraph{Prompt C}
\begin{quote}
\ttfamily
You are given a sequence of images showing a scene unfolding over
time. Examine each consecutive pair of frames: does the transition from
one image to the next always make physical sense? Two adjacent frames may
have been swapped, causing one transition to look reversed or impossible.
Reply with only `yes' if you find such a swap, or `no' if all transitions
look natural.
\end{quote}

\paragraph{Prompt D}
\begin{quote}
\ttfamily
Look at these images in order. Have any two neighboring images
been swapped? Reply with only `yes' or `no'.
\end{quote}

\paragraph{Prompt E}
\begin{quote}
\ttfamily
You are given a sequence of images. Go through them one by one
from the first to the last. For each consecutive pair, ask yourself:
could this transition happen naturally - does the second image
physically follow from the first? If any single transition looks reversed
or impossible, two adjacent frames have been swapped. Reply with only
`yes' if you find such a transition, or `no' if every step forward in
the sequence looks natural.
\end{quote}

\subsection{Scene Description Templates}
\label{appendix:scene_description}

Scene descriptions are appended as plain text after the image tokens in the user message. For CLEVRER and CRAFT they are taken verbatim from the datasets' event annotations; for DriveLM from the nuScenes scene-level metadata; for MTL-AQA they are generated from the structured dive annotations (rotation type, body position, somersault and twist count, armstand flag). Examples:

\paragraph{CLEVRER}
\begin{quote}
Brown cube will collide with blue cylinder, and blue cylinder will collide with gray ball.
\end{quote}
\begin{quote}
Yellow ball will collide with green ball, green ball will collide with yellow cylinder, and green ball will collide with gray cube.
\end{quote}
\begin{quote}
Green cube will collide with gray ball, green cube will collide with blue cylinder, and gray ball will collide with gray ball.
\end{quote}
\begin{quote}
Yellow cube will collide with blue ball, and blue ball will collide with yellow ball.
\end{quote}

\paragraph{CRAFT}
\begin{quote}
Large blue triangle will enter the basket.
\end{quote}
\begin{quote}
Large blue triangle will collide with large gray triangle, small cyan triangle will collide with large purple triangle, large blue triangle will enter the basket, and large gray triangle will collide with large purple triangle.
\end{quote}
\begin{quote}
Small cyan triangle will collide with large purple triangle, large blue triangle will enter the basket, and large purple triangle will enter the basket.
\end{quote}
\begin{quote}
Large blue triangle will collide with large gray triangle, and large blue triangle will enter the basket.
\end{quote}

\paragraph{DriveLM}
\begin{quote}
The ego vehicle halted at the intersection with traffic lights, awaiting the passage of preceding vehicles before
  veering to the right.
\end{quote}
\begin{quote}
The ego vehicle proceeded straight after making a left turn at the current intersection, and is about to make a right turn.
\end{quote}
\begin{quote}
The ego vehicle traverses along the current roadway, encountering construction on both the left and right sides.
\end{quote}
\begin{quote}
The ego vehicle is traveling along the current road and is about to pass through the traffic light intersection.
\end{quote}

\paragraph{MTL-AQA}
\begin{quote}
A forward dive with 2.5 somersaults in pike position.
\end{quote}
\begin{quote}
An armstand forward dive with 2 somersaults and 1.5 twists in tuck position.
\end{quote}
\begin{quote}
A reverse dive with 3.5 somersaults in straight position.
\end{quote}
\begin{quote}
An inward dive with 2.5 somersaults in a straight position.
\end{quote}

\section{Dataset Details}
\label{appendix_dataset_details}

\begin{figure*}
    \centering
    \includegraphics[width=1\linewidth]{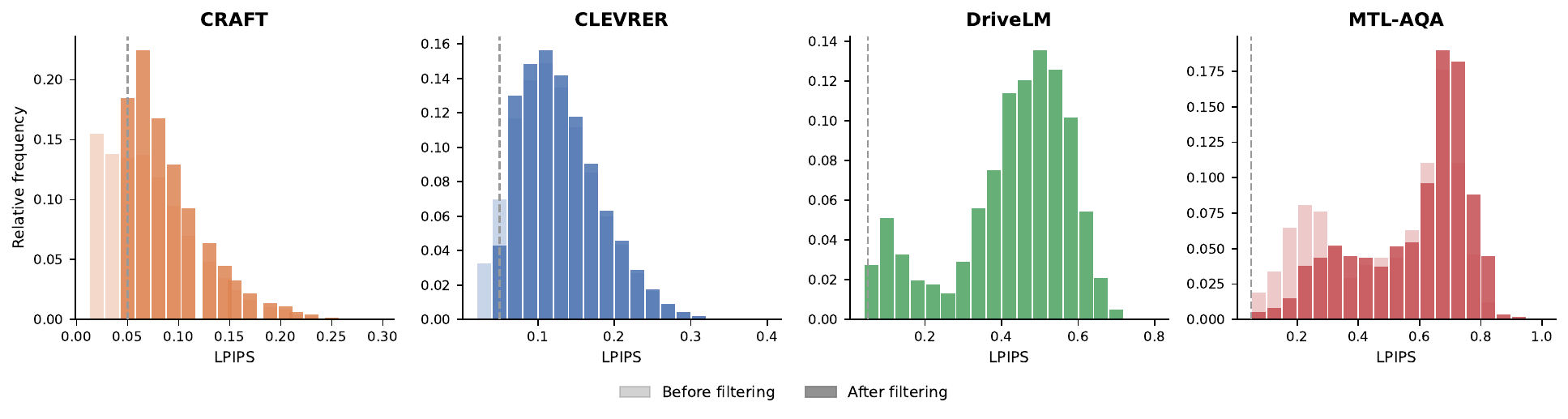}
    \caption{LPIPS distributions of consecutive frame pairs before and after filtering. Filtering removes nearly identical frame pairs while preserving the overall distribution of perceptual differences. Synthetic datasets (CRAFT and CLEVRER) contain more visually similar consecutive frames than the real-world datasets due to their largely static backgrounds.}
    \label{fig:lpips-filtering}
\end{figure*}

Table~\ref{tab:dataset_stats} summarizes the characteristics of the four datasets before and after sampling and filtering, including the number of sequences, average sequence length, and perceptual similarity (LPIPS). Figure~\ref{fig:lpips-filtering} illustrates the LPIPS distributions before and after filtering, highlighting the removal of nearly identical consecutive frames used to construct the benchmark.

\paragraph{Licensing} CLEVRER is released under CC0. CRAFT is released
under CC BY 4.0. DriveLM's annotations are released under CC
BY-NC-SA~4.0. MTL-AQA is used consistent with standard academic reuse of an established public benchmark\,--\,the original release does not specify an explicit license.

\begin{table*}[ht]
    \centering
    \setlength{\tabcolsep}{6pt}
    \begin{tabular}{llcccc}
        \toprule
        \textbf{Dataset} & \textbf{Content} & \textbf{Video} & \textbf{Sequences} & \textbf{Avg.\ frames} & \textbf{Mean\ LPIPS} \\
        \midrule
        CRAFT & Abstract 2D objects, collisions &  \checkmark & $1{,}983 \to 1{,}959$ & $16.0 \to 9.0$ & $0.053 \to 0.093$ \\
        CLEVRER & Abstract 3D objects, collisions & \checkmark & $5{,}000 \to 4{,}997$ & $8.0 \to 7.5$  & $0.123 \to 0.131$ \\
        DriveLM & Driving, egocentric & - & $696$ & $5.9$ & $0.428$ \\
        MTL-AQA & Olympic competitive diving & \checkmark
        & $353 \to 338$ & $8.4 \to 4.5$  & $0.492 \to 0.584$ \\
        \bottomrule
    \end{tabular}
    \caption{Dataset statistics before and after the sampling and filtering pipeline. Arrows indicate the changes in the number of sequences, average sequence length, and mean LPIPS after preprocessing. The 1,959 filtered CRAFT sequences are partitioned into 858 main-evaluation sequences (4–8 frames) and 1,101 CRAFT-long sequences (9–16 frames).}
    \label{tab:dataset_stats}
\end{table*}

\section{Human Study Details}
\label{appendix:human_study}

\begin{figure*}[t!]
    \centering
    \includegraphics[width=1\textwidth]{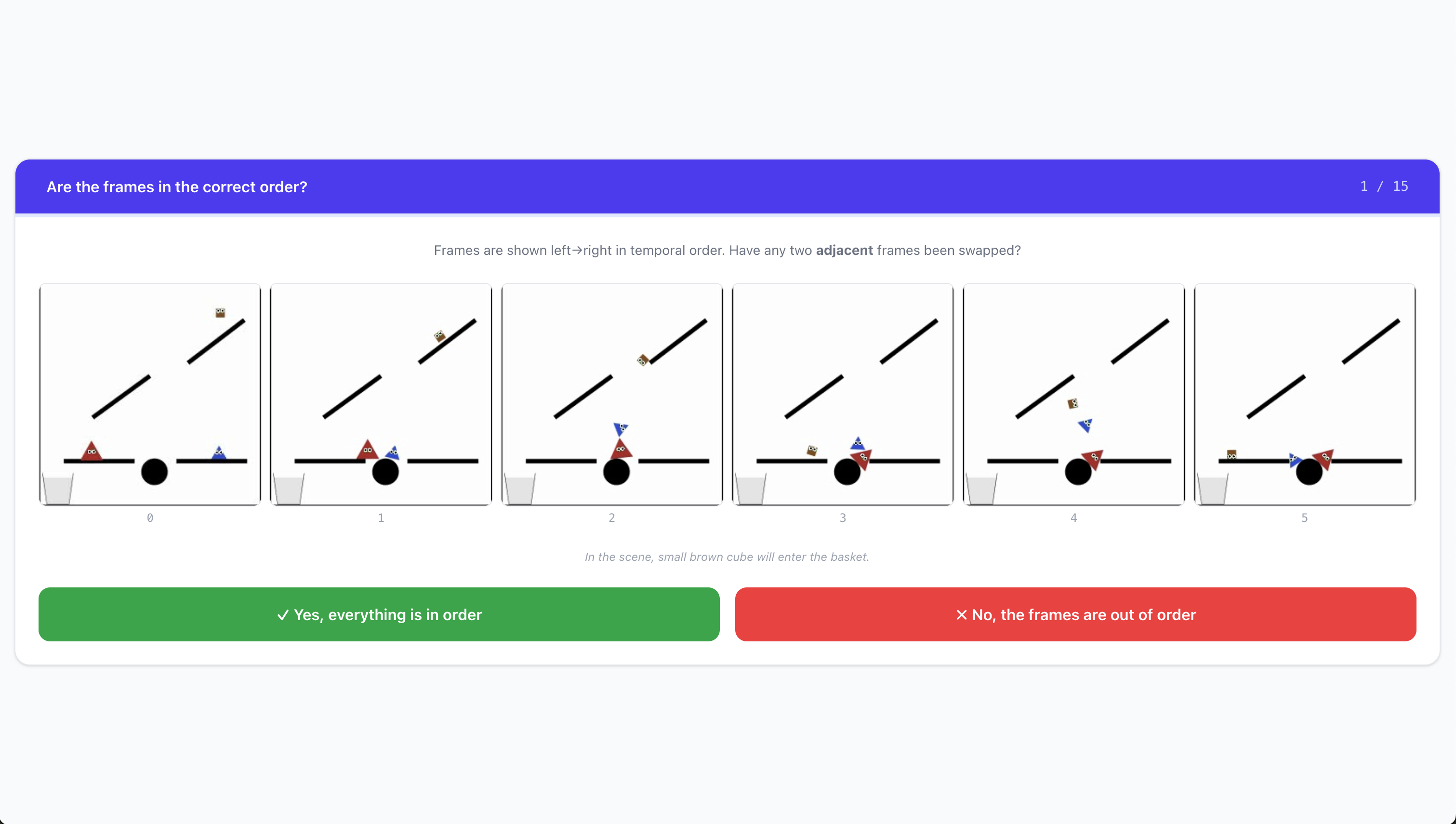}
    \caption{User interface for the temporal anomaly detection task. Participants viewed an image sequence and indicated whether the frames were presented in the correct temporal order. Scene descriptions were provided below the sequence to match the model evaluation setting.}
    \label{fig:annotation-interface}
\end{figure*}

\begin{figure*}[t!]
    \centering
    \includegraphics[width=1\textwidth]{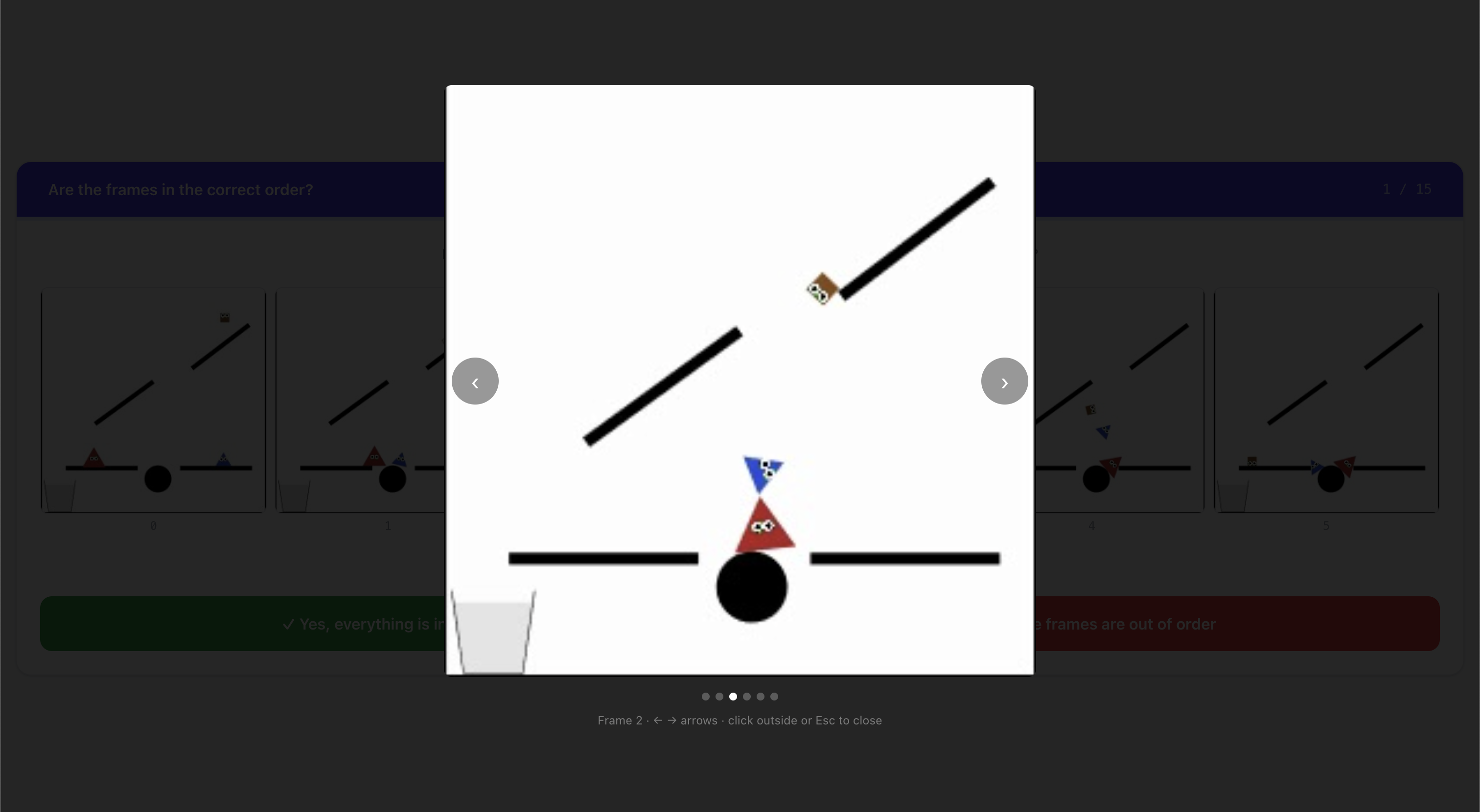}
    \caption{Close-up of the image viewer. Viewer was beneficial for annotators to notice finer details, \,--\,it allows for quicker swapping between the images to emulate a more video-like experience.}
    \label{fig:annotation-interface-viewer}
\end{figure*}

\paragraph{Interface}
We developed a custom annotation interface that presents participants with an image sequence alongside its scene description and guides them through either the detection or localization task. Examples of the interface are shown in Figure~\ref{fig:annotation-interface}. Participants can also preview each sequence using an interactive viewer navigable with arrow keys, illustrated in Figure~\ref{fig:annotation-interface-viewer}. Prior to annotation, participants complete an onboarding flow covering the annotation guidelines, including visual examples and a video walkthrough of the full annotation process. The platform is integrated with Prolific, which was used to recruit and compensate participants.

\subsection{Annotation Guidelines}

Each experimental condition was accompanied by a dedicated instruction page describing the video domain and the annotation task. The instructions shown to participants for the CRAFT dataset are presented below.

\paragraph{Introduction}
We are conducting a study about whether AI systems can understand time in videos. Concretely: if you show AI a sequence of images from a video, can it tell if the images are in the right order? To answer this, we also need to know how well humans perform the same task.

You will be given a series of four to eight \textbf{snapshots} from a video. The \textbf{goal} is to \textbf{evaluate} how well humans can detect if a swap happened in the sequence. This data will then be used to compare against how well AI performs. No other data than the annotations is collected (no personal data, location data, etc.).

\paragraph{Your Task}
You will interact with a dataset of simple 2D shapes (circles, squares, triangles in different colors and sizes) moving, colliding, rolling down ramps, and sometimes falling into a basket.

You will be given a sequence where two \textbf{consecutive} images \textbf{may have been} swapped. Inspect the sequence and use the buttons to select whether the sequence was modified with a swap, or is still in the correct order.

\textbf{Tip:} Click any image to open a full-size viewer and use the \textbf{$\leftarrow$ $\rightarrow$} arrow keys (or the on-screen buttons) to flip between frames one at a time --- this makes it much easier to spot a swap.

The full workflow is illustrated in the accompanying video.

The whole process should take around 15 minutes across 15 sequences. Thank you for your participation!

\subsection{Respondents}

We used Prolific\footnote{https://www.prolific.com/} to recruit and compensate study participants. We recruited 24 participants (3 per dataset and task configuration), all holding at least an undergraduate degree. The sample was slightly unbalanced towards male participants (16 male, 8 female), with a mean age of 35.2 years. Participants were geographically diverse, spanning 16 countries of residence. Each participant completed a single experimental condition and was compensated at an effective mean hourly rate exceeding \pounds8.00/hour, above Prolific's minimum recommended reward of \pounds6.00 per hour. The mean completion time was 9.8 minutes for 15 image sequences.
\subsection{Control Mechanisms}

To check for potentially invalid or rushed study submissions, we have incorporated 3 simple sequences into each participant's task to serve as attention checks. Participants who failed two or all three attention checks were then rejected. The attention check sequences were manually selected to be as easy as possible and then excluded from the human study results.

\section{Additional Results}
\label{appendix:additional-results}

\begin{figure*}[t]
    \centering
    \includegraphics[width=1\linewidth]{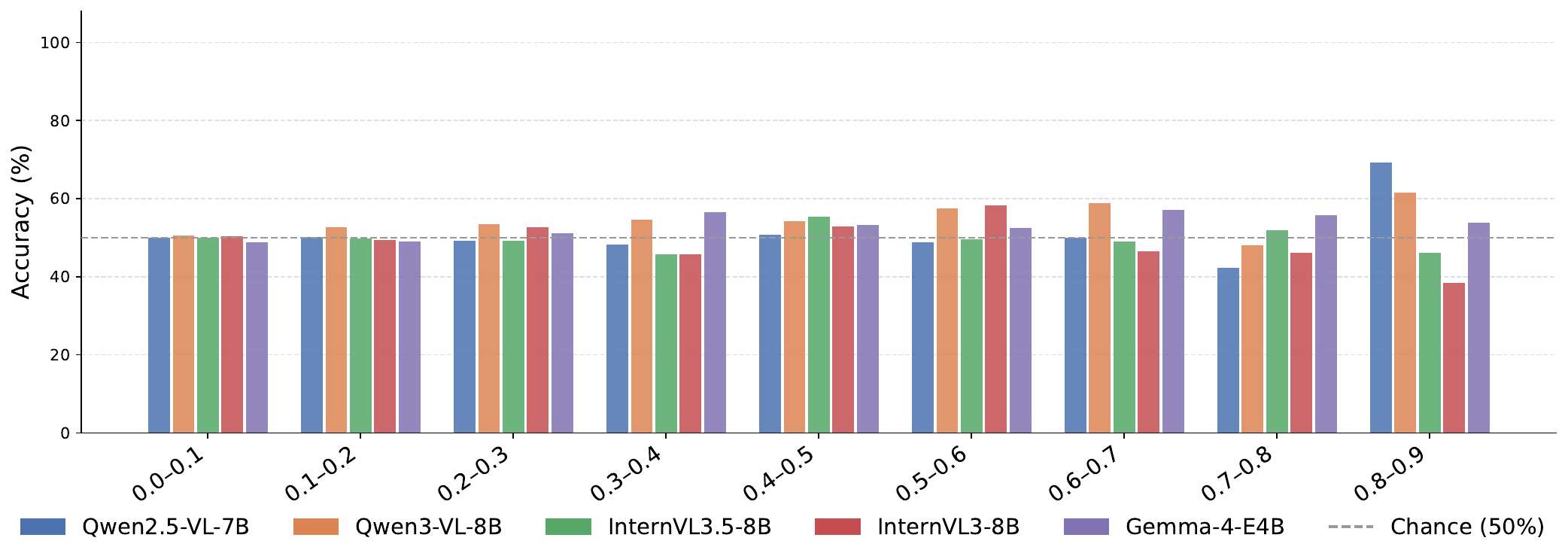}
    \caption{Temporal anomaly detection accuracy by LPIPS distance for all evaluated models. Accuracy remains close to chance across most perceptual similarity bins, indicating that larger visual differences between swapped frames do not consistently improve performance. The apparent increase in the final bin ($0.8$--$0.9$) is likely due to the small number of samples ($n=13$).}
    \label{fig:lpips-bins}
\end{figure*}

\begin{figure*}[t]
    \centering
    \includegraphics[width=1\linewidth]{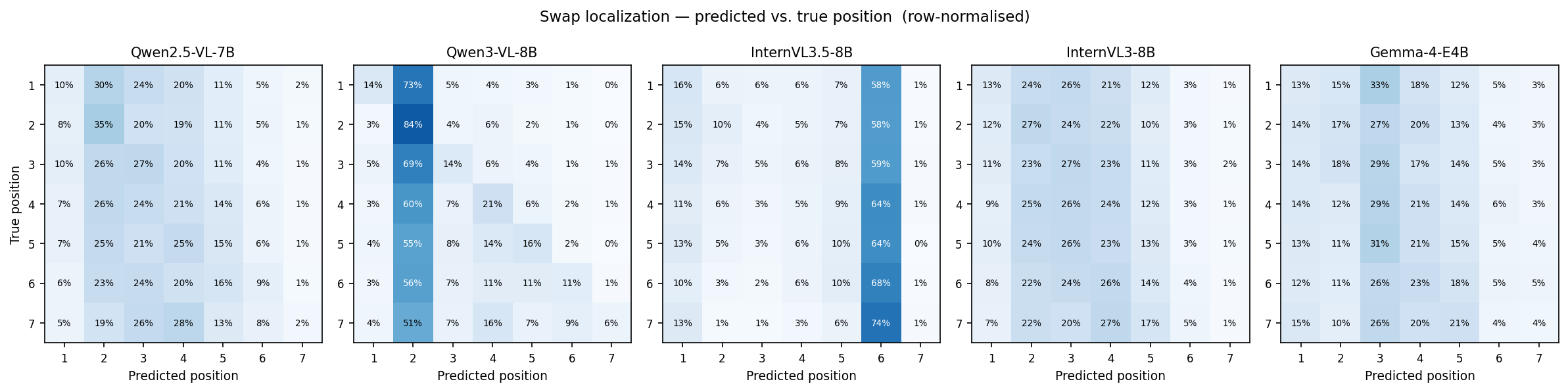}
    \caption{Temporal localization confusion matrices for all evaluated models, averaged across datasets and row-normalized. Two distinct error patterns emerge. Qwen3-VL and InternVL3.5 exhibit a strong bias toward predicting specific swap positions, whereas Gemma-4, Qwen2.5-VL, and InternVL3 produce more uniformly distributed predictions, consistent with near-random guessing.}
    \label{fig:localize-heatmap}
\end{figure*}

Table~\ref{tab:no_scene_desc_delta} reports the per-model change in accuracy after removing scene descriptions. Figure~\ref{fig:lpips-bins} presents temporal anomaly detection accuracy across LPIPS distance bins for all evaluated models. Figure~\ref{fig:localize-heatmap} shows the remaining localization confusion matrices, complementing the analysis in Section \ref{sec:analysis}.

\paragraph{Can Temporal Anomaly Detection be Learned?} To investigate whether the temporal consistency limitations observed in pretrained VLMs can be mitigated through targeted supervision, we conduct a preliminary fine-tuning experiment using Qwen3-VL-2B. The model is fine-tuned on the temporal anomaly detection task using the training split (80\%) of the CLEVRER dataset and evaluated on the held-out validation split (20\%), as well as on the remaining datasets (CRAFT, DriveLM, and MTL-AQA) without further adaptation. Fine-tuning is performed using LoRA~\cite{hu2022lora} through the ms-swift~\cite{zhao2024swiftascalablelightweightinfrastructure}. The training is then run for 3 epochs with learning rate $1\times10^{-4}$, effective batch size 16, and max sequence length 4096 tokens. We report the final checkpoint (step 714). All training is conducted on an NVIDIA A100 GPU with seed 42.

\begin{table}[h]
    \centering
    \setlength{\tabcolsep}{4pt}
    \begin{tabular}{l cccc}
    \toprule
     & CL & CR & DR & MT \\
    \midrule
    QW3-2B (0-shot) & 49.6 &  50.5 & 50.0 & 44.1  \\
    QW3-2B (FT) & \textbf{96.5} & \textbf{63.6} & \textbf{52.7} & \textbf{68.1} \\
    \quad $\Delta$ & +46.9 & +13.1 & +2.7 & +24.0 \\ 
    \midrule
    QW3-8B (0-shot) & 51.9 & 53.3 & 51.9 & 57.4 \\
    \bottomrule
    \end{tabular}
    \caption{Temporal anomaly detection accuracy (\%) of Qwen3-VL-2B (QW3-2B) before and after fine-tuning on the CLEVRER training split. Fine-tuning consistently improves performance on the held-out CLEVRER split and transfers to the unseen CRAFT, DriveLM, and MTL-AQA datasets. Zero-shot Qwen3-VL-8B (QW3-8B) is shown for reference.}
    \label{tab:finetuning}
\end{table}

Table~\ref{tab:finetuning} reports the results. Fine-tuning improves temporal anomaly detection on the CLEVRER test split as well as on the unseen CRAFT, DriveLM, and MTL-AQA datasets. Notably, the fine-tuned 2B model matches or exceeds the zero-shot performance of the larger Qwen3-VL-8B model evaluated in the main paper. While these experiments are preliminary and limited to a single model family, they suggest that at least part of the observed limitation is attributable to the supervision available during training rather than model scale alone. Targeted temporal supervision improves temporal anomaly detection while generalizing across unseen datasets spanning both synthetic and real-world domains. We leave a controlled study across model scales and model families for future work. FT: fine-tuned.

\newcommand{\ci}[1]{{\fontsize{7}{8}\selectfont\color{black!55}#1}}
\begin{table*}[t!]
    \centering
    \setlength{\tabcolsep}{1.5pt}
    \small
    \begin{tabular*}{\textwidth}{@{\extracolsep{\fill}}lcccccc@{}}
    \toprule
     & Random & Qwen2.5-VL-7B & Qwen3-VL-8B & Gemma-4-E4B & InternVL3-8B & InternVL3.5-8B \\
    \midrule
    \multicolumn{7}{@{}l}{\textit{Temporal Detect}} \\
    \midrule
    \quad CLEVRER & 50.0 & 50.5\,\ci{[49.1,\,51.9]} & \textbf{51.9}\,\ci{[50.5,\,53.3]} & 49.2\,\ci{[47.9,\,50.6]} & 50.1\,\ci{[48.7,\,51.5]} & 49.6\,\ci{[48.2,\,51.0]} \\
    \quad CRAFT & 50.0 & 47.6\,\ci{[44.2,\,50.9]} & \textbf{53.3}\,\ci{[49.9,\,56.6]} & 48.0\,\ci{[44.7,\,51.4]} & 49.1\,\ci{[45.7,\,52.4]} & 49.9\,\ci{[46.5,\,53.2]} \\
    \quad DriveLM & 50.0 & 50.7\,\ci{[47.0,\,54.4]} & 51.9\,\ci{[48.2,\,55.6]} & \textbf{53.9}\,\ci{[50.2,\,57.6]} & 53.2\,\ci{[49.4,\,56.8]} & 49.9\,\ci{[46.2,\,53.6]} \\
    \quad MTL-AQA & 50.0 & 45.3\,\ci{[40.0,\,50.6]} & \textbf{57.4}\,\ci{[52.1,\,62.6]} & 56.2\,\ci{[50.9,\,61.4]} & 47.9\,\ci{[42.7,\,53.2]} & 50.6\,\ci{[45.3,\,55.9]} \\
    \midrule
    \multicolumn{7}{@{}l}{\textit{Temporal Localize}} \\
    \midrule
    \quad CLEVRER & 4.3 & 13.1\,\ci{[12.2,\,14.0]} & \textbf{25.6}\,\ci{[24.4,\,26.8]} & 15.0\,\ci{[14.1,\,16.0]} & 14.3\,\ci{[13.3,\,15.3]} & 15.4\,\ci{[14.4,\,16.4]} \\
    \quad CRAFT & 5.5 & 14.3\,\ci{[12.1,\,16.8]} & \textbf{24.5}\,\ci{[21.7,\,27.5]} & 14.7\,\ci{[12.5,\,17.2]} & 17.7\,\ci{[15.3,\,20.4]} & 18.2\,\ci{[15.7,\,20.9]} \\
    \quad DriveLM & 8.7 & 14.8\,\ci{[12.4,\,17.6]} & \textbf{24.7}\,\ci{[21.7,\,28.1]} & 22.3\,\ci{[19.3,\,25.5]} &17.8\,\ci{[15.2,\,20.8]} & 9.8\,\ci{[7.8,\,12.2]} \\
    \quad MTL-AQA & 13.2 & 29.9\,\ci{[25.3,\,35.0]} &\textbf{43.2}\,\ci{[38.0,\,48.5]} & 29.9\,\ci{[25.3,\,35.0]} & 21.6\,\ci{[17.5,\,26.3]} & 33.1\,\ci{[28.3,\,38.3]} \\
    \midrule
    \multicolumn{7}{@{}l}{\textit{Frame Detect}} \\
    \midrule
    \quad CLEVRER & 50.0 & 66.9\,\ci{[65.6,\,68.2]} &\textbf{93.3}\,\ci{[92.6,\,94.0]} & 77.2\,\ci{[76.1,\,78.4]} & 58.1\,\ci{[56.7,\,59.4]} & 66.9\,\ci{[65.6,\,68.2]} \\
    \quad CRAFT & 50.0 & 67.9\,\ci{[64.8,\,71.0]} & \textbf{98.1}\,\ci{[97.0,\,98.8]} & 80.5\,\ci{[77.8,\,83.0]} &77.7\,\ci{[74.8,\,80.4]} & 66.7\,\ci{[63.4,\,69.7]} \\
    \quad DriveLM & 50.0 & 60.3\,\ci{[56.7,\,63.9]} &\textbf{99.6}\,\ci{[98.7,\,99.9]} & 82.0\,\ci{[79.0,\,84.7]} & 68.4\,\ci{[64.8,\,71.7]} & 69.7\,\ci{[66.2,\,73.0]} \\
    \quad MTL-AQA & 50.0 & 71.6\,\ci{[66.6,\,76.1]} & \textbf{94.1}\,\ci{[91.0,\,96.1]} & 76.6\,\ci{[71.8,\,80.8]} &76.6\,\ci{[71.8,\,80.8]} & 77.2\,\ci{[72.5,\,81.4]} \\
    \midrule
    \multicolumn{7}{@{}l}{\textit{Frame Localize}} \\
    \midrule
    \quad CLEVRER & 13.5 & 42.0\,\ci{[40.6,\,43.4]} & \textbf{98.9}\,\ci{[98.5,\,99.1]} & 33.6\,\ci{[32.3,\,35.0]} &30.9\,\ci{[29.7,\,32.2]} & 18.6\,\ci{[17.5,\,19.7]} \\
    \quad CRAFT & 14.9 & 60.5\,\ci{[57.2,\,63.7]} &\textbf{96.3}\,\ci{[94.8,\,97.3]} & 46.5\,\ci{[43.2,\,49.8]} & 52.9\,\ci{[49.6,\,56.2]} & 22.4\,\ci{[19.7,\,25.3]} \\
    \quad DriveLM & 18.1 & 45.8\,\ci{[42.2,\,49.5]} & \textbf{92.7}\,\ci{[90.5,\,94.4]} & 61.9\,\ci{[58.3,\,65.5]} &33.9\,\ci{[30.5,\,37.5]} & 16.4\,\ci{[13.8,\,19.3]} \\
    \quad MTL-AQA & 22.4 & 73.1\,\ci{[68.1,\,77.5]} &\textbf{99.4}\,\ci{[97.9,\,99.8]} & 88.2\,\ci{[84.3,\,91.2]} & 65.7\,\ci{[60.5,\,70.5]} & 29.6\,\ci{[25.0,\,34.7]} \\
    \bottomrule
    \end{tabular*}
    \caption{Detection and localization accuracy (\%) across datasets (same runs as Table~\ref{tab:detailed_results}), with 95\% Wilson score confidence intervals. For temporal detect, most intervals include the 50\% chance level.}
    \label{tab:detailed_results_ci}
\end{table*}

\begin{table*}[t!]
    \centering
    \setlength{\tabcolsep}{4pt}
    \small
    \begin{tabular}{l cccc cccc cccc cccc}
    \toprule
     & \multicolumn{4}{c}{Temporal Detect} & \multicolumn{4}{c}{Temporal Localize} & \multicolumn{4}{c}{Frame Detect} & \multicolumn{4}{c}{Frame Localize} \\
    \cmidrule(lr){2-5} \cmidrule(lr){6-9} \cmidrule(lr){10-13} \cmidrule(lr){14-17}
     & CL & CR & DR & MT & CL & CR & DR & MT & CL & CR & DR & MT & CL & CR & DR & MT \\
    \midrule
    QW2.5 & \cellcolor{red!5}-0.5 & \cellcolor{green!15}+4.4 & \cellcolor{red!10}-1.7 & \cellcolor{green!20}+7.7 & \cellcolor{green!10}+1.8 & \cellcolor{green!13}+3.1 & \cellcolor{green!14}+3.6 & \cellcolor{green!16}+4.7 & \cellcolor{green!12}+2.6 & \cellcolor{green!11}+2.2 & \cellcolor{green!13}+2.9 & \cellcolor{red!16}-4.7 & \cellcolor{green!43}+33.9 & \cellcolor{green!32}+19.3 & \cellcolor{green!36}+23.3 & \cellcolor{green!27}+13.0 \\
    QW3 & \cellcolor{green!20}+7.0 & \cellcolor{green!11}+2.3 & \cellcolor{red!7}-1.0 & \cellcolor{green!31}+17.2 & \cellcolor{green!16}+4.7 & \cellcolor{green!11}+2.1 & \cellcolor{red!2}-0.1 & \cellcolor{green!6}+0.6 & \cellcolor{green!17}+5.3 & \cellcolor{green!6}+0.6 & \cellcolor{red!7}-0.9 & \cellcolor{red!15}-4.1 & \cellcolor{green!7}+1.0 & \cellcolor{green!12}+2.6 & \cellcolor{green!8}+1.1 & +0.0 \\
    GEM4 & \cellcolor{green!6}+0.7 & \cellcolor{green!11}+2.2 & \cellcolor{red!17}-5.0 & \cellcolor{red!7}-0.9 & \cellcolor{red!3}-0.2 & \cellcolor{green!3}+0.2 & \cellcolor{green!11}+2.3 & \cellcolor{red!11}-2.1 & \cellcolor{green!20}+7.5 & \cellcolor{green!10}+1.7 & \cellcolor{green!17}+5.6 & \cellcolor{green!9}+1.5 & \cellcolor{green!12}+2.5 & \cellcolor{red!12}-2.7 & \cellcolor{green!18}+5.7 & \cellcolor{green!6}+0.6 \\
    IVL3 & \cellcolor{green!2}+0.1 & \cellcolor{green!12}+2.6 & \cellcolor{red!6}-0.6 & \cellcolor{green!22}+8.9 & \cellcolor{green!8}+1.3 & \cellcolor{red!13}-2.9 & \cellcolor{green!10}+1.7 & \cellcolor{green!11}+2.4 & \cellcolor{green!30}+16.0 & \cellcolor{green!22}+8.7 & \cellcolor{green!25}+11.8 & \cellcolor{green!7}+0.9 & \cellcolor{green!29}+15.9 & \cellcolor{green!26}+12.7 & \cellcolor{green!39}+28.3 & \cellcolor{green!33}+20.1 \\
    IVL3.5 & \cellcolor{green!5}+0.5 & \cellcolor{green!6}+0.6 & \cellcolor{red!10}-2.0 & \cellcolor{green!24}+10.9 & \cellcolor{green!12}+2.5 & \cellcolor{green!10}+2.0 & \cellcolor{green!24}+10.9 & \cellcolor{green!13}+3.3 & \cellcolor{green!23}+9.8 & \cellcolor{green!22}+9.0 & \cellcolor{green!25}+11.9 & \cellcolor{green!20}+7.4 & \cellcolor{green!46}+39.6 & \cellcolor{green!43}+34.6 & \cellcolor{green!55}+55.0 & \cellcolor{green!60}+66.0 \\
    \bottomrule
    \end{tabular}
    \caption{Change in accuracy (\%) after removing scene descriptions. Positive values ({\color{green!70!black}green}) indicate improved performance without scene descriptions, while negative values ({\color{red}red}) indicate degradation. Removing scene descriptions has the largest effect on frame-level localization, whereas temporal anomaly performance changes only modestly.}
    \label{tab:no_scene_desc_delta}
\end{table*}

\begin{table*}[t!]
    \centering
    \setlength{\tabcolsep}{3pt}
    \small
    \begin{tabular}{l rrrrl rrrrl}
    \toprule
     & \multicolumn{5}{c}{Temporal Detect} & \multicolumn{5}{c}{Temporal Localize} \\
    \cmidrule(lr){2-6} \cmidrule(lr){7-11}
    Model & $n$ & Human & Model & Gap & $p$ & $n$ & Human & Model & Gap & $p$ \\
    \midrule
    Qwen2.5-VL-7B & 167 & 83.8 & 51.5 & +32.3 & $1.1\times10^{-8}$ *** & 143 & 86.7 & 22.4 & +64.3 & $6.4\times10^{-23}$ *** \\
    Qwen3-VL-8B & 167 & 83.8 & 58.1 & +25.7 & $6.1\times10^{-6}$ *** & 143 & 86.7 & 29.4 & +57.3 & $8.8\times10^{-19}$ *** \\
    Gemma-4-E4B & 167 & 83.8 & 49.7 & +34.1 & $1.9\times10^{-8}$ *** & 143 & 86.7 & 26.6 & +60.1 & $6.6\times10^{-23}$ *** \\
    InternVL3-8B & 167 & 83.8 & 53.3 & +30.5 & $9.1\times10^{-9}$ *** & 143 & 86.7 & 14.7 & +72.0 & $6.7\times10^{-27}$ *** \\
    InternVL3.5-8B & 167 & 83.8 & 51.5 & +32.3 & $1.8\times10^{-10}$ *** & 143 & 86.7 & 25.2 & +61.5 & $8.8\times10^{-22}$ *** \\
    \bottomrule
    \end{tabular}
    \caption{Human vs.\ model accuracy (\%), McNemar exact test (Bonferroni corrected), averaged across all four datasets per model. Each model is compared to the human annotators on the identical samples it was evaluated on.}
    \label{tab:mcnemar_pooled}
\end{table*}

\begin{figure*}[p]
	\centering
	
	\begin{subfigure}[b]{\linewidth}
		\centering
		\includegraphics[width=\linewidth]{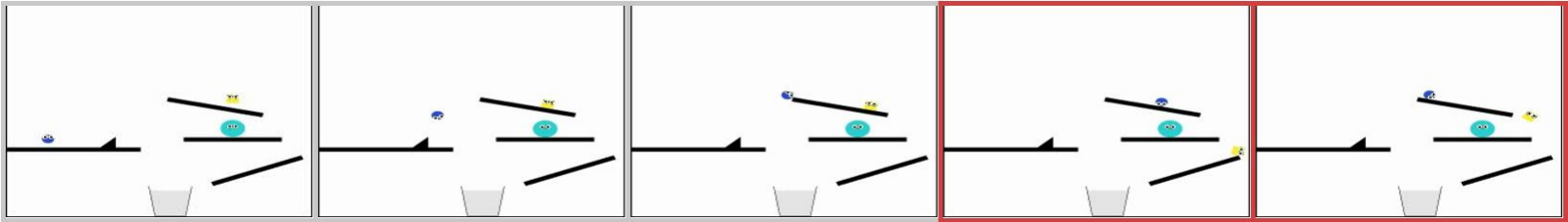}
	\end{subfigure}
	
	\smallskip
	
	\begin{subfigure}[b]{\linewidth}
		\centering
		\includegraphics[width=0.8\linewidth]{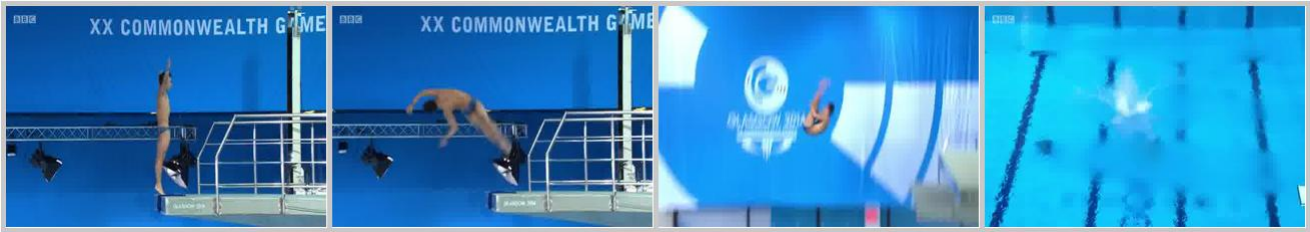}
	\end{subfigure}
	
	\smallskip
	
	\begin{subfigure}[b]{\linewidth}
		\centering
		\includegraphics[width=\linewidth]{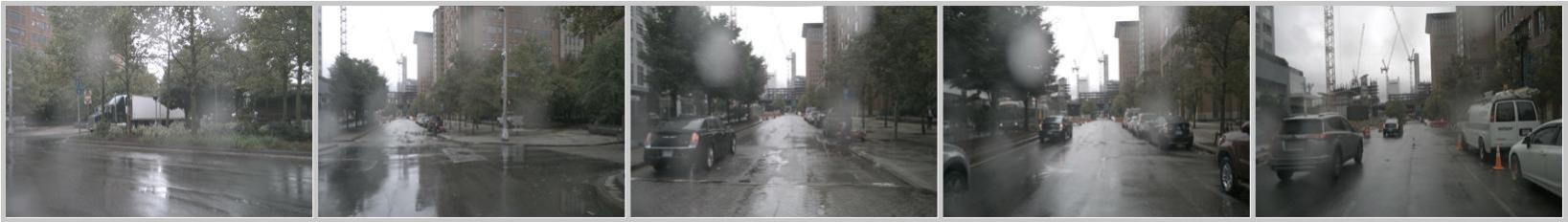}
	\end{subfigure}
	
	\smallskip
	
	\begin{subfigure}[b]{\linewidth}
		\centering
		\includegraphics[width=\linewidth]{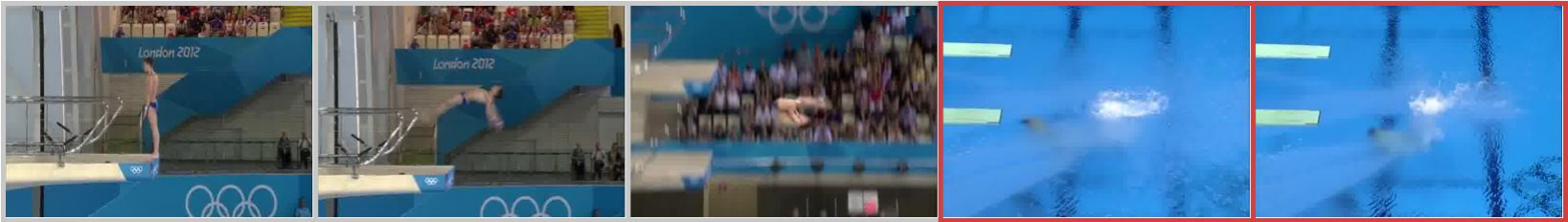}
	\end{subfigure}
	
	\caption{%
		Qualitative examples illustrating success and failure modes of both humans and VLMs.
		Red borders mark the swapped frame pair. From top to bottom:
		\textbf{(1)}~\textit{Both correct\,--\,swap present} (CRAFT, detect):
		the pair at positions~3--4 is out of order, human and all five models
		correctly identify the swap.
		\textbf{(2)}~\textit{Both correct\,--\,no swap} (MTL-AQA, detect):
		the sequence is unmodified, human and all models correctly report no anomaly.
		\textbf{(3)}~\textit{LLMs correct, human wrong} (Drive-LM, detect):
		the sequence is unmodified, but the human incorrectly reports a swap,
		likely misled by rainy conditions and a camera turn between frames;
		all five models answer correctly.
		\textbf{(4)}~\textit{Both wrong} (MTL-AQA, localize):
		the swap at positions~3--4 (two nearly identical underwater frames)
		is missed by both the human annotator and all five models,
		who instead point to earlier positions in the sequence.
	}
	\label{fig:quadrant-examples}
\end{figure*}

\end{document}